\documentclass{article}

\usepackage{microtype}
\usepackage{graphicx}
\usepackage{subfigure}
\usepackage{booktabs} 
\usepackage{amsmath} 

\usepackage{graphicx}
\usepackage{tabularx}
\usepackage{enumerate}
\usepackage{multirow}
\usepackage{listings}
\usepackage{xcolor}
\usepackage{comment}
\usepackage{xurl}

\definecolor{codegreen}{rgb}{0,0.6,0}
\definecolor{codegray}{rgb}{0.5,0.5,0.5}
\definecolor{codepurple}{rgb}{0.58,0,0.82}
\definecolor{backcolour}{rgb}{0.95,0.95,0.92}

\lstdefinestyle{mystyle}{
	backgroundcolor=\color{backcolour},   
	commentstyle=\color{codegreen},
	keywordstyle=\color{magenta},
	numberstyle=\tiny\color{codegray},
	stringstyle=\color{codepurple},
	basicstyle=\ttfamily\footnotesize,
	breakatwhitespace=false,         
	breaklines=true,                 
	captionpos=b,                    
	keepspaces=true,                 
	numbers=left,                    
	numbersep=5pt,                  
	showspaces=false,                
	showstringspaces=false,
	showtabs=false,                  
	tabsize=2
}

\usepackage{hyperref} 

\usepackage[accepted]{icml2019}

\begin{document}

\twocolumn[
\title{CrystalCandle: A User-Facing Model Explainer for Narrative Explanations}
\author{Jilei Yang, LinkedIn Corporation, jlyang@linkedin.com \\ Diana Negoescu, LinkedIn Corporation, dnegoescu@linkedin.com \\ Parvez Ahammad, LinkedIn Corporation, pahammad@linkedin.com}
\date{\vspace{-0.2in}}
\maketitle

\begin{abstract}
Predictive machine learning models often lack interpretability, resulting in low trust from model end users despite having high predictive performance. While many model interpretation approaches return top important features to help interpret model predictions, these top features may not be well-organized or intuitive to end users, which limits model adoption rates. In this paper, we propose CrystalCandle, a user-facing model explainer that creates user-digestible interpretations and insights reflecting the rationale behind model predictions. CrystalCandle builds an end-to-end pipeline from machine learning platforms to end user platforms, and provides users with an interface for implementing model interpretation approaches and for customizing narrative insights. CrystalCandle is a platform consisting of four components: Model Importer, Model Interpreter, Narrative Generator, and Narrative Exporter. We describe these components, and then demonstrate the effectiveness of CrystalCandle through use cases at LinkedIn. Quantitative performance analyses indicate that CrystalCandle's narrative insights lead to lifts in adoption rates of predictive model recommendations, as well as to increases in downstream key metrics such as revenue when compared to previous approaches, while qualitative analyses indicate positive feedback from end users.
\\

\end{abstract}
\vspace{0.4in}
]
\clearpage

\section{Introduction}\label{sec:introduction}
Predictive machine learning models are widely used in a variety of areas in industry. For example, in sales and marketing, predictive models can help determine which potential customers are likely to purchase a product, and in healthcare, they can assist clinicians in detecting the risks of certain diseases. Complex predictive models such as random forest, gradient boosted trees, and deep neural networks can produce more accurate predictions than simple models such as linear regression and decision trees, and are therefore preferred in many use cases where prediction accuracy is of utmost importance. However, one important challenge is explaining model predictions to end users who are experts in their domains, using application-specific platforms and language. Previous literature points out that users can be reluctant to use the predictive models if they do not understand how and why these models make predictions \cite{martens2014explaining}, \cite{moeyersoms2016explaining}. Therefore, building a user-facing model explainer that provides model interpretation and feature reasoning becomes crucial for engendering trust in prediction results and creating meaningful insights based on them.

Unfortunately, most complex predictive models with high predictive performance are intrinsically opaque, causing difficulties in intuitive interpretations. Even though some models output a list of globally important features to interpret the overall model prediction, usually no interpretations at individual sample level are produced. For example, in sales prediction, it may be that for customer A, \texttt{browsing time} is the most important feature whereas for customer B, \texttt{discount} is the most important. A sales team may strategize different customers individually by learning each customer’s own top features. Therefore, developing a user-facing model explainer which provides feature reasoning at individual sample level is of critical need \cite{ribeiro2016should}.

There exist several state-of-the-art model interpretation approaches that enable sample-level feature reasoning, e.g., LIME \cite{ribeiro2016should}, KernelSHAP \cite{lundberg2017unified}, and TreeSHAP \cite{lundberg2020local}. These approaches produce feature importance scores for each sample, indicating how much each feature has contributed to the model prediction. A typical example of model prediction and interpretation results using LIME for a jobs upsell model in LinkedIn sales prediction is shown in Table \ref{table:model_output_narrative}. Here, a random forest model predicts how likely each LinkedIn customer is to purchase more job slot products at contract renewal by using over 100 features covering areas such as job slots usage, job seeker activity, and company level attributes.

\begin{table*}[t]
	\caption{Model prediction \& interpretation result (left panel) and narrative insights (right panel).}\label{table:model_output_narrative}
	\begin{tabularx}{\textwidth}{|X|X|}
		\hline
		\multicolumn{1}{|c|}{\textbf{Model Prediction \& Interpretation (Non-Intuitive)}}
		& \multicolumn{1}{c|}{\textbf{Narrative Insights (User-Friendly)}} \\
		\hline
		Propensity score: 0.85 (Top 2\%)\newline Top important features (with importance score): \begin{itemize} \item paid\_job\_s4: 0.030 \item job\_view\_s4: 0.013 \item hire\_cntr\_s3: 0.011 \item conn\_cmp\_s4: 0.009 \item $\cdots$ \end{itemize} & This account is extremely likely to upsell. Its upsell likelihood is larger than 98\% of all accounts, which is driven by: \begin{itemize} \item Paid job posts changed from 10 to 15 (+50\%) in the last month. \item Views per job changed from 200 to 300 (+50\%) in the last month. \item $\cdots$ \end{itemize}
		\\ \hline
	\end{tabularx}
\end{table*}

The left panel of Table \ref{table:model_output_narrative} displays the model outputs with the interpretation results for a specific customer in jobs upsell prediction. Here, even though we have conducted sample-level feature reasoning by providing top important feature lists, there still exist several challenges when surfacing model interpretation results to end users such as sales teams:
\begin{enumerate}
	\item
	Feature names in top important feature list may not be easily understood by sales teams. For example, feature names such as \texttt{hire\_cntr\_s3} and \texttt{conn\_cmp\_s4} may be too abstract for sales team to extract meaningful information.
	\item
	Top important feature lists may not be well-organized. For example, some features are closely related to each other and can be grouped, some features contain too many details for sales team to digest, and some features are not very meaningful to sales team and can be removed.
	\item
	Top important feature lists may not be easily integrated into sales management platforms which use sales-friendly language, resulting in low adoption rates by sales teams on predictive modeling intelligence.
\end{enumerate}

\noindent A desired resolution is to convert these non-intuitive model interpretation results into user-digestible narrative insights as shown in the right panel of Table \ref{table:model_output_narrative}.

Two common approaches for creating narratives in the literature are generation-based and template-based \cite{zhang2018explainable}. The former relies on neural-network-based natural language generation techniques, and can generate narrative explanations in an automatic way to effectively save human effort. However, the generation-based approach usually requires a large training data set, which may not exist in many use cases -- for example, in sales prediction, our main use case, existing narratives for sales recommendations are very limited, as sales teams usually do not write logs of how they reached decisions about their customers. On the other hand, template-based approaches can achieve a much higher generalizability than generation-based approaches, as they do not rely on such training data. For this reason, we choose the template-based approach for creating narrative explanations.

In this paper, we propose a template-based, user-facing model explainer - CrystalCandle, which aims to create user-understandable interpretation and insights, and to reflect the rationale behind machine learning models. In contrast with sample-level model interpretation approaches, CrystalCandle has added follow-up steps to convert non-intuitive model interpretation results into user-digestible narrative insights. In contrast with other approaches, such as rule-based narrative generation systems, CrystalCandle is more scalable as it has leveraged machine learning models to conduct automatic narrative ranking.

To our knowledge, CrystalCandle is the first user-facing model explainer employed in industry that provides a user-friendly interface to conduct model interpretation and narrative customization. Besides sales and marketing, CrystalCandle is applicable in a variety of use cases in industry. For example, in healthcare, CrystalCandle can help identify top signals in disease risk prediction, and then translate them into clinical reports which are digestible to clinicians. In credit card applications, narratives containing main reasons why applicants get rejected can be generated by CrystalCandle from credit risk prediction models, and then surfaced to credit card applicants for their reference.

The rest of the paper is organized as follows. Section \ref{sec:related_work} lists related work in the area of model interpretation and narrative generation; Section \ref{sec:intellige_design} describes the design of CrystalCandle; Section \ref{sec:use_case} presents use cases of CrystalCandle at LinkedIn, including performance evaluation results; Section \ref{sec:limitations} points out some limitations of CrystalCandle and discusses future directions; and Section \ref{sec:conclusion} concludes our work.

\section{Related Work}\label{sec:related_work}

Model interpretation approaches that focus on sample-level feature reasoning have been widely explored in recent years. Examples include Shapley Value \cite{strumbelj2010efficient}\cite{vstrumbelj2014explaining}, Local Gradient \cite{baehrens2010explain}, Integrated Gradient \cite{sundararajan2017axiomatic}, Quantitative Input Influence (QII) \cite{datta2016algorithmic}, Leave-One-Covariate-Out (LOCO) \cite{lei2018distribution}, LIME \cite{ribeiro2016should}, KernelSHAP \cite{lundberg2017unified}, and TreeSHAP \cite{lundberg2020local}. Moreover, many model interpretation platforms have also been developed to facilitate the implementation of these approaches in a unified way, e.g., Microsoft InterpretML \cite{nori2019interpretml} and Machine Learning Interpretability (MLI) in H2O Driverless AI \cite{hall2019introduction}. All these model interpretation approaches and platforms can easily suffer from one challenge when interpretation results are presented to end users: feature importance scores in tabular/bar-chart format may not be very intuitive, resulting in low adoption rates.

To overcome this limitation, user-digestible narrative-based model interpretations have been proposed \cite{reiter2019natural}\cite{baaj2019some}. Two common approaches for such interpretations are generation-based and template-based. Examples of neural network generation-based approaches include synthesizing explanations triggered by the word “because” \cite{antol2015vqa}\cite{hendricks2016generating}, leveraging LSTM to generate explanation sentences \cite{costa2018automatic}, creating tips for Yelp restaurants based on GRU \cite{li2017neural}, and developing a multi-task recommendation model which performs rating prediction and recommendation explanation simultaneously \cite{lu2018like}. However, generation-based approaches highly depend on the quality and quantity of training data, thus are less generalizable than template-based approaches.

Recent work on creating narrative explanations via template-based approaches includes imputing the predefined narrative templates with the most important features to explain the recommendation models \cite{zhang2014explicit}\cite{wang2018explainable}\cite{tao2019fact}. In \cite{biran2014justification}, a Java package provides narrative justifications for logistic/linear regression models, \cite{calegari2019interpretable} propose a way to generate narrative explanations using logical knowledge translated from a decision tree model, and \cite{sovrano2019difference}  introduce a rule-based explainer for a GDPR automated decision which applies to explainable models. However, all these aforementioned templated-based approaches are only applicable to a subset of machine learning models, and can easily fail when facing a more complex model such as a random forest. Moreover, the templates used in these approaches are predefined, with limited variations, and as a result, the generated narratives can become repetitive and hard to customize. In CrystalCandle, we overcome these limitations by implementing model-agnostic interpretation approaches which apply to arbitrary predictive machine learning models, and by providing a user-friendly interface that allows customizing an unlimited number of narrative templates.

\section{CrystalCandle Design}\label{sec:intellige_design}

\subsection{Overview}\label{subsec:overview}

We propose CrystalCandle as a self-service platform for user-facing explanation. CrystalCandle demystifies the outputs of predictive models by assigning feature importance scores, and converts non-intuitive model predictions and top important features into user-understandable narratives. This enables end users to obtain insights into model predictions, and to build trust in model recommendations.

CrystalCandle is designed to support all the commonly-used black-box supervised machine learning models, including but not limited to support vector machines, bagging, random forests, gradient boosted trees, and deep neural networks.

Several challenges existed in the design and deployment of CrystalCandle:
\begin{enumerate}
	\item
	How to consume outputs from a range of machine learning platforms implementing machine learning models?
	\item
	How to enable flexibility in choosing model interpretation approaches for different use cases?
	\item
	How to efficiently generate template-based narratives while allowing narrative customization?
	\item
	How to produce narratives compatible with a range of end user platforms?
\end{enumerate}

\begin{figure*}[t]
	\centering
	\includegraphics[width=0.85\textwidth]{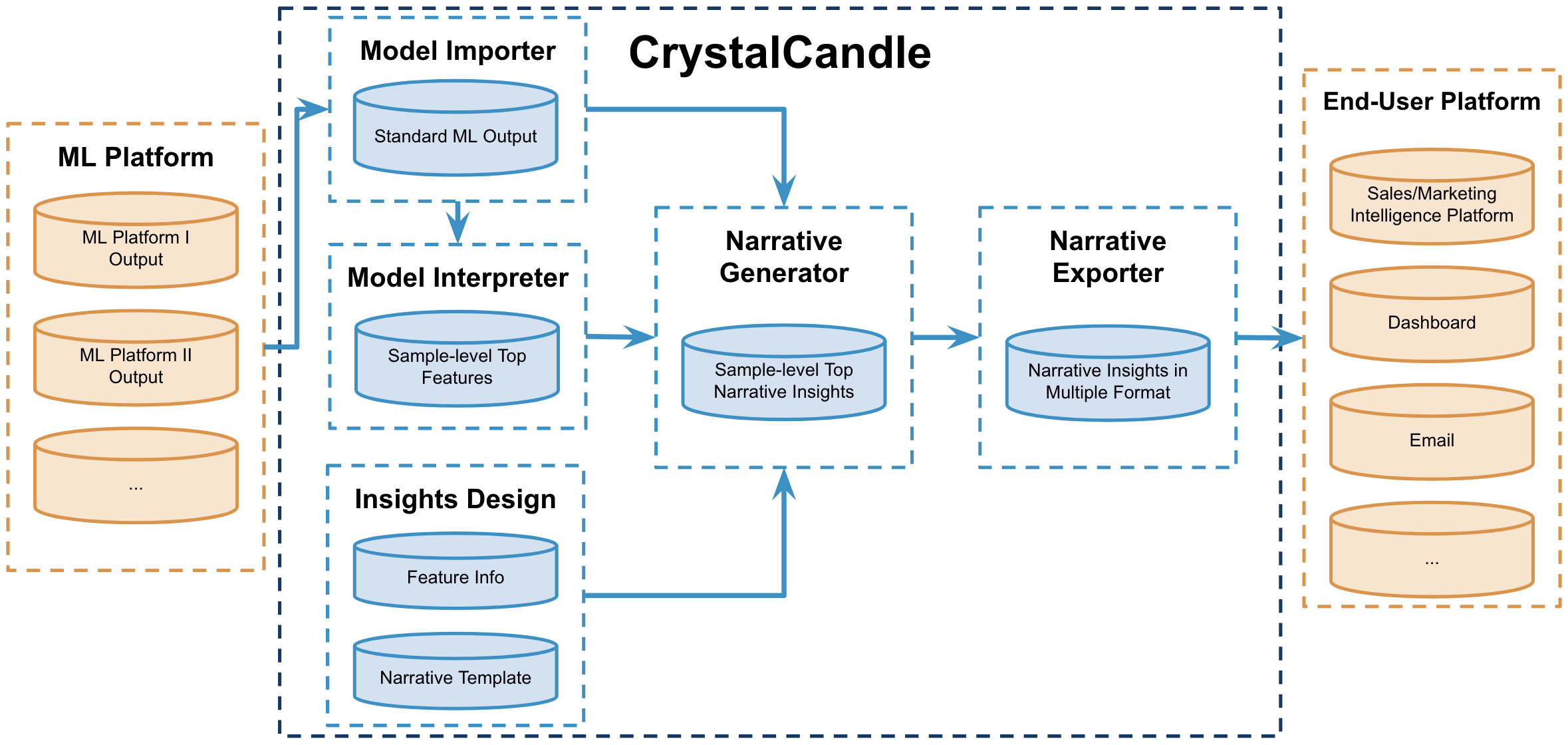}
	\caption{CrystalCandle Components.} \label{fig:intellige_components}
\end{figure*}

\noindent To address the above challenges, we designed CrystalCandle as a flexible platform consisting of four components: Model Importer, Model Interpreter, Narrative Generator and Narrative Exporter. These four components resolved the above challenges in sequential order. Figure \ref{fig:intellige_components} shows these four components:
\begin{enumerate}
	\item
	\textbf{Model Importer:} Consumes model output from major machine learning platforms and transforms it into standardized machine learning model output.
	\item
	\textbf{Model Interpreter:} Implements a collection of model interpretation approaches to process the standardized machine learning model output, and produces sample-level top important feature lists.
	\item
	\textbf{Narrative Generator:} Creates user-digestible narratives via a template-based approach, based on the standardized machine learning model output, and additional feature information and narrative templates provided by CrystalCandle users; Selects top narratives by using top important feature list for each sample.
	\item
	\textbf{Narrative Exporter:} Surfaces sample-level top narratives onto major end user platforms with necessary format adjustments.
\end{enumerate}

\noindent In the following sections, we introduce these four components in more detail.

\subsection{Model Importer}\label{subsec:model_importer}

As we move toward machine learning platforms such as ProML from LinkedIn,\footnote{https://engineering.linkedin.com/blog/2019/01/scaling-machine-learning-productivity-at-linkedin} AutoML from Google,\footnote{https://cloud.google.com/automl} and Create ML from Apple,\footnote{https://developer.apple.com/machine-learning/create-ml} it is very likely that different platforms produce model outputs in very different formats, resulting in low efficiency of developing model explainers specifically for each platform. A natural resolution is to first convert these model outputs into standardized format. This leads to the development of Model Importer.

The Model Importer takes the model output from a set of machine learning platforms as its input, and produces standardized machine learning model output, which will be used in the following Model Interpreter and Narrative Generator. For use cases at LinkedIn, the set of machine learning platforms includes ProML and other internal platforms built by data science teams. A typical standardized machine learning model output consists of feature vectors and prediction scores of all the samples, and optionally the predictive model itself with a unified interface (i.e., the interface should take standardized input (feature vectors) and produce standardized output (prediction scores)). We set the predictive model to be optional, since the following Model Interpreter can sometimes work well even without access to the original model, and the Narrative Generator does not depend on the original model.

\subsection{Model Interpreter}

Model Interpreter is the second component of CrystalCandle, aiming to reveal insights behind machine learning model recommendations. It takes the output of Model Importer as its input, and produces sample-level top important feature lists by calculating feature importance scores for each sample, which are then conveyed to the Narrative Generator as one of its inputs.

The Model Interpreter consists of a collection of model interpretation approaches with a unified input format (i.e., standardized machine learning model output) and a unified output format (i.e., sample-level top important feature lists). The collection of model interpretation approaches includes state-of-the-art methods that produce sample-level feature importance scores, e.g., LIME \cite{ribeiro2016should}, KernelSHAP \& DeepSHAP \cite{lundberg2017unified}, and TreeSHAP \cite{lundberg2020local}. The Model Interpreter need not have the model itself accessible  -- the model may be trained in a separate system/device or there exist privacy/security concerns. If that is the case, high-performing model interpretation approaches, such as K-LIME, are available \cite{hall2019introduction}.

CrystalCandle users have the flexibility to choose the appropriate interpretation approach in their use cases: For example, if the input machine learning platform implements a restricted set of machine learning algorithms (e.g., only tree-based algorithms or neural-network-based algorithms), then CrystalCandle users can choose model-specific interpretation approaches such as TreeSHAP and DeepSHAP as they are usually more computationally-efficient than model-agnostic ones; On the contrary, if the original platform keeps a large set of candidate algorithms, then model-agnostic interpretation approaches such as LIME and KernelSHAP are recommended.

\subsection{Narrative Generator}\label{subsec:narrative_generator}

Narrative Generator is the key innovative part of CrystalCandle, as it generates human-understandable narratives for interpreting model predictions. Its input consists of the outputs from Model Importer and Model Interpreter, as well as Insights Design, which in turn consists of a Feature Info File and Narrative Templates (shown in Figure \ref{fig:intellige_components}, details provided in Section \ref{subsubsecrion:feature_hierarchy}-\ref{subsubsection:narrative_concatenation}). The design of Narrative Generator is challenging: Information solely from the model itself such as feature name, feature value, and feature importance score may not be comprehensive enough for narrative construction. Additional information such as feature descriptions and narrative templates are needed as well.

To address the above challenge, we propose the Narrative Generator, with the goal of minimizing the human effort in narrative generation while keeping the flexibility in narrative customization. Many heavy tasks, such as feature value extraction, template imputation, and narrative ranking are handled inside the Narrative Generator in an automated way.

To enable narrative customization, we introduce Insights Design, an additional input to Narrative Generator provided by CrystalCandle users. Insights Design contains information from domain knowledge owners which cannot be directly extracted from the model itself, but is essential to narrative construction. Insights Design has two components: Feature Info File and Narrative Templates. The Feature Info File contains additional information for each model feature, including feature hierarchy information, detailed feature descriptions, and narrative template imputation rules. The Narrative Templates file contains a collection of templates for imputing appropriate feature values.

In the following sections, we describe key features of Narrative Generator in detail.

\subsubsection{Four-Layer Feature Hierarchy}\label{subsubsecrion:feature_hierarchy}

An intuitive way to understand feature meaning and the intrinsic relationship between different features is to construct a hierarchical structure for these features. In CrystalCandle, we propose a four-layer feature hierarchy, where the original features are set to be the first layer. This feature hierarchy is manually constructed with the help of model owners, and is specified by Feature Info File in Insights Design. In practice, feature correlation analysis can also help in finding the appropriate hierarchical structure by providing feature grouping information.

Table \ref{table:feature_hierarchy} shows a sample feature hierarchy for 8 selected features from the jobs upsell model introduced in Section \ref{sec:introduction}. There are four hierarchical layers for features in Table \ref{table:feature_hierarchy}:
\begin{table*}[t]
	\caption{Feature hierarchy for selected features from jobs upsell model.}\label{table:feature_hierarchy}
	\centering
	\begin{tabular}{|c|c|c|c|}
		\hline
		\textbf{Original-Feature} & \textbf{Super-Feature} & \textbf{Ultra-Feature} & \textbf{Category} \\
		\hline
		job\_qty & job slots & job slots & product booking \\
		\hline
		job\_dprice\_usd & job slots & job slots & product booking \\
		\hline
		job\_view\_s3 & views per job & job view & product performance \\
		\hline
		job\_view\_s4 & views per job & job view & product performance \\
		\hline
		job\_viewer\_s3 & viewers per job & job view & product performance \\
		\hline
		job\_viewer\_s4 & viewers per job & job view & product performance \\
		\hline
		job\_applicant\_s3 & applicants per job & job applicant & product performance \\
		\hline
		job\_applicant\_s4 & applicants per job & job applicant & product performance \\
		\hline
	\end{tabular}
\end{table*}

\begin{itemize}
	\item
	\textbf{Original-feature layer (1st layer):} This layer contains all the original features used in the model.
	\item
	\textbf{Super-feature layer (2nd layer):} This layer is used to group closely related original-features for narrative construction. For each super-feature, one narrative will be constructed, and every original-feature under this super-feature will have the opportunity to appear in this narrative. Another important functionality of this layer is to conduct feature name explanation. As we can see from Table \ref{table:feature_hierarchy}, the super-feature names are much more understandable than the original-feature names, and they themselves are likely to appear in the narrative as well. For example, one narrative will be constructed for the super-feature \texttt{views per job}, where the values of the original-features \texttt{job\_view\_s3} and \texttt{job\_view\_s4} as well as the super feature name \texttt{views per job} are likely to be incorporated into the narrative. We discuss the details of how to construct narratives in Section \ref{subsubsection:narrative_template_imputation}.
	\item
	\textbf{Ultra-feature layer (3rd layer):} This layer is used to identify and filter out super-features with (almost) duplicated information to end users. For example, in Table \ref{table:feature_hierarchy}, both super-features \texttt{views per job} and \texttt{viewers per job} are under ultra-feature \texttt{job view} as they are very likely to contain overlapped information. If we show one narrative of \texttt{views per job} to end users, then the marginal benefit to show the other narrative of \texttt{viewers per job} will be low. We discuss how to filter out super-features under one ultra-feature in Section \ref{subsubsection:narrative_ranking}.
	\item
	\textbf{Category layer (4th layer):} This layer is used to group relevant narratives and concatenate them into paragraphs. For example, in Table \ref{table:feature_hierarchy}, both ultra-features \texttt{job view} and \texttt{job applicant} belong to category \texttt{product performance}, and their corresponding narratives can then be concatenated into one paragraph by using conjunction phrases. We discuss the details of narrative concatenation in Section \ref{subsubsection:narrative_concatenation}.
\end{itemize}

\noindent We note that when the Narrative Generator was initially designed, it only consumed the original-feature and the super-feature layers as its input, as these two layers seemed necessary and sufficient in narrative generation. However, as more and more requests for narrative deduplication and narrative concatenation came in from CrystalCandle users, we decided to add the ultra-feature and the category layers, in order to realize these two additional functionalities. In practice, CrystalCandle users can choose whether to specify the ultra-feature and category layers based on their own use cases. For example, specifying the ultra-feature layer is recommended if the generated narratives contain too much redundant information, and specifying the category layer is recommended if end users prefer reading paragraphs. If CrystalCandle users find it unnecessary to specify either of these two layers, they can simply set them the same as the super-feature layer to reduce preparation effort.

\subsubsection{Narrative Template Imputation}\label{subsubsection:narrative_template_imputation}

An important prerequisite for narrative construction is building the narrative templates in Insights Design. Narrative templates are manually constructed with the help of model owners, and then translated into appropriate code for imputing feature values in Narrative Generator. An example of how to translate narrative templates into code can be found in Section \ref{subsec:translate_template} in the Appendix. 

Table \ref{table:narrative_templates} shows sample narrative templates for the jobs upsell use case. Here, we refer to \texttt{value\_change} as an “insight type”: Each super-feature corresponds to one insight type, which determines the specific narrative template to use.  \texttt{prev\_value} and \texttt{current\_value} in the template \texttt{value\_change} are “insight items”: each original-feature under one super-feature corresponds to one insight item, which determines the position to impute the original-feature value into the narrative template. For example, as shown in Table \ref{table:insight_type}, the original-features \texttt{job\_view\_s3} and \texttt{job\_view\_s4} under the super-feature \texttt{views per job} correspond to insight items \texttt{prev\_value} and \texttt{current\_value} respectively, where \texttt{prev\_value} and \texttt{current\_value} can be identified as two positions in the template \texttt{value\_change} in Table \ref{table:narrative_templates}. \texttt{percent\_change} is an example of an “extra insight item” whose value may not be directly extracted from original-features but can be derived by extra calculations on the existing insight items. For example, here \texttt{percent\_change = (current\_value-prev\_value)/prev\_value*100}. We note that the appearance of the original-features within a given narrative is solely determined by the design of the narrative template, rather than the importance scores of the original-features -- we discuss the usage of these importance scores in Section \ref{subsubsection:narrative_ranking}. We also mention that, by default, all the original-features under one super-feature will appear in its corresponding narrative.

The introduction of “insight type” and “insight item” enables the reusability of narrative templates. For example, both super-features \texttt{views per job} and \texttt{applicants per job} share the same insight type, and thus their narratives are constructed based on the same template. Moreover, “insight item” enables the construction of sample-specific narratives, as each sample has its own feature values to be imputed.

In addition to Narrative Templates, we are now able to specify the complete version of Feature Info File within Insights Design: The Feature Info File consists of all the columns in Table \ref{table:feature_hierarchy} and \ref{table:insight_type}: Original-Feature, Super-Feature, Ultra-Feature, Category, Insight Type, and Insight Item, which contains essential feature information for narrative construction. Three additional columns: Insight Threshold, Insight Weight, and Source, can also be incorporated as optional columns to make the narrative generation process more customizable. The detailed introduction of these three columns can be found in Section \ref{subsec:additional_column} in the Appendix.

\begin{table*}[h]
	\caption{Narrative templates for interpreting jobs upsell model.}\label{table:narrative_templates}
	\centering
	\begin{tabularx}{0.6\textwidth}{|>{\hsize=0.25\hsize}X|>{\hsize=0.75\hsize}X|}
		\hline
		\multicolumn{1}{|c|}{\textbf{Insight Type}}
		& \multicolumn{1}{c|}{\textbf{Narrative Template}} \\
		\hline
		\centering quantity & Purchased \{quantity\_num\} \{super\_name\} for \$\{total\_price\}. \\
		\hline
		\centering value\_change & \{super\_name\} changed from \{prev\_value\} to \{current\_value\} (\{percent\_change\}\%) in the last month. \\
		\hline
	\end{tabularx}
	\\ ~\\
	\caption{Insight type and insight item for selected features from jobs upsell model.}\label{table:insight_type}
	\centering
	\begin{tabular}{|c|c|c|c|}
		\hline
		\textbf{Original-Feature} & \textbf{Super-Feature} & \textbf{Insight Type} & \textbf{Insight Item} \\
		\hline
		job\_qty & job slots &quantity & quantity\_num \\
		\hline
		job\_dprice\_usd & job slots &quantity & total\_price \\
		\hline
		job\_view\_s3 & views per job & value\_change & prev\_value \\
		\hline
		job\_view\_s4 & views per job & value\_change & current\_value \\
		\hline
		job\_viewer\_s3 & viewers per job & value\_change & prev\_value \\
		\hline
		job\_viewer\_s4 & viewers per job & value\_change & current\_value \\
		\hline
		job\_applicant\_s3 & applicants per job & value\_change & prev\_value \\
		\hline
		job\_applicant\_s4 & applicants per job & value\_change & current\_value \\
		\hline
	\end{tabular}
\end{table*}

\subsubsection{Narrative Ranking}\label{subsubsection:narrative_ranking}

By introducing the feature info file and narrative templates, we are able to construct a collection of narratives for each sample. However, too many narratives may overwhelm end users, so instead we aim to present them with a few selected ones which show the strongest signals to support the model recommendations. 

To select the most important narratives in a scalable way, we leverage the sample-level top important feature list created from Model Interpreter to rank all the narratives, and then present the end users with the top ones. To this end, we introduce the narrative importance score, which reflects how large the contribution of each narrative is to the model prediction. We set the score to be the maximum importance score among all the original-features under the super-feature corresponding to the narrative. The intuition behind this setting is that one narrative is important as long as at least one of its corresponding original-features is important.

We conduct narrative deduplication by retaining only the top $K$ narratives with the largest narrative importance scores under each ultra-feature (Recall that the ultra-feature is in a higher hierarchy of the super-feature, which groups those super-features sharing the overlapped information). 
We can set $K=1$ to make the generated narratives most concise. One example of conducting narrative ranking and deduplication in jobs upsell use case can be found in Section \ref{subsec:narrative_ranking_example} in the Appendix (Table \ref{table:narrative_ranking}).

Finally, we point out that narrative ranking has inherited a good property from feature ranking in Model Interpreter: narrative ranking is sample-specific. A narrative with the same content, e.g., \texttt{views per job}, can be ranked as No. 1 for customer A but No. 5 for customer B, indicating its different contributions in supporting the recommendations for different customers.

\subsubsection{Narrative Concatenation}\label{subsubsection:narrative_concatenation}

Narrative concatenation is enabled by the category layer of the four-layer feature hierarchy, where relevant narratives under the same category can be concatenated as a paragraph rather than a bullet-point list. The major goal is to make the narratives better-organized so that the narratives focusing on different aspects of the sample will not be mixed. In CrystalCandle, narrative concatenation is optional.

We conduct narrative concatenation by using conjunction phrases such as “and”, “moreover” and “what’s more”. For example, for narratives corresponding to super-features \texttt{views per job} and \texttt{applicants per job}, the paragraph after narrative concatenation is “Views per job changed ..., and applicants per job changed ...”. We also introduce paragraph importance score to rank these paragraphs. Similar to narrative importance score, the paragraph importance score is determined as the largest narrative importance score among all the narratives incorporated in the paragraph.

\subsubsection{Narrative Generator Design}

\begin{figure*}[t]
	\centering
	\includegraphics[width=0.85\textwidth]{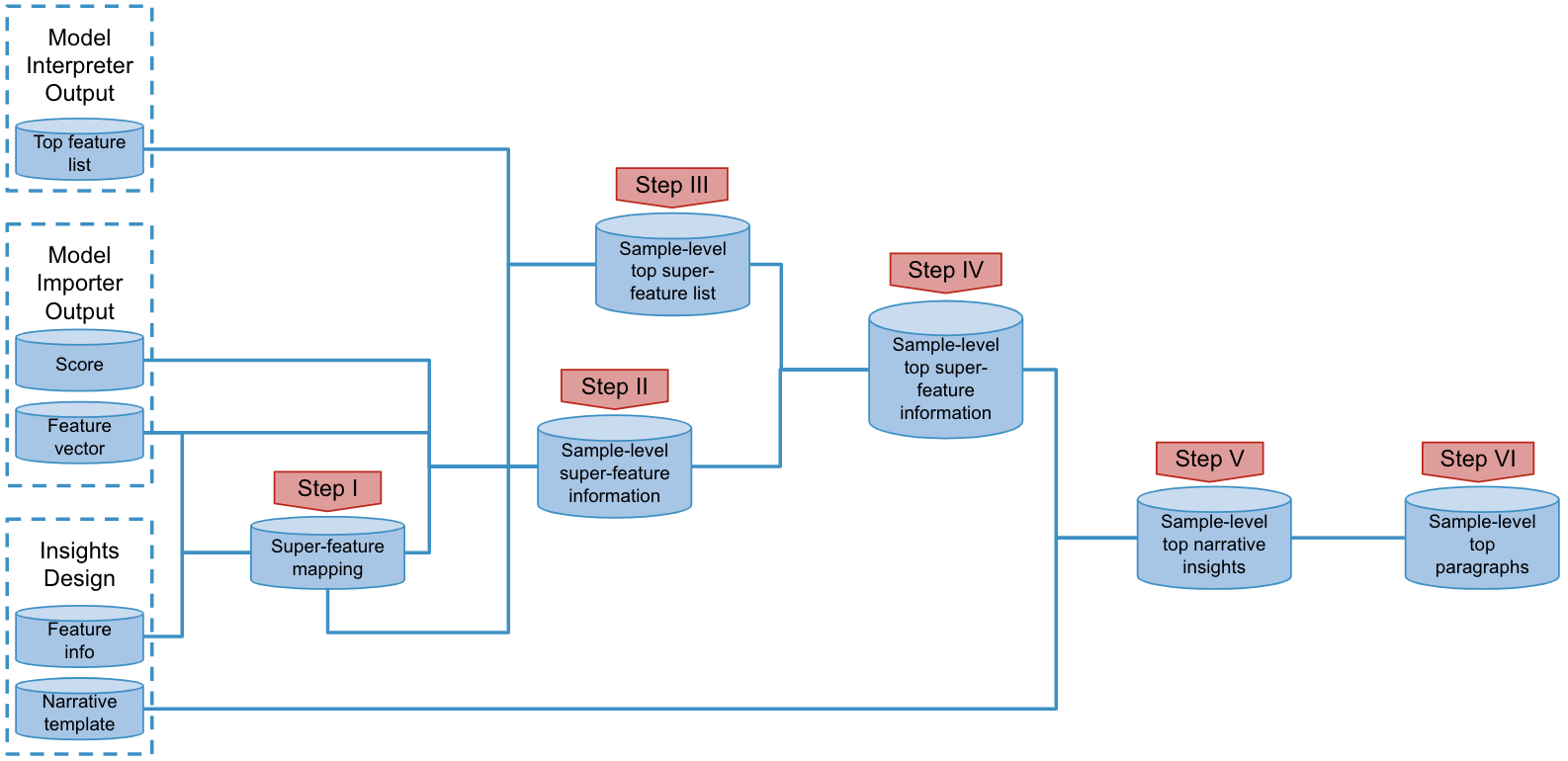}
	\caption{Narrative Generator - Major Steps.} \label{fig:narrative_generator}
\end{figure*}

We now describe the design of Narrative Generator. Figure \ref{fig:narrative_generator} shows the six major steps in Narrative Generator:

\begin{enumerate}[I]
	\item
	Construct super-feature mapping based on feature info file and feature vectors: For each super-feature, this mapping records its corresponding original-feature ids (i.e., positions in feature vector), ultra-feature, category, insight type and insight items.
	\item
	Collect information of all super-features for each sample: For each super-feature in one sample, we extract its corresponding original-feature values from feature vectors according to super-feature mapping from Step I.
	\item
	Obtain top super-feature list for each sample: Based on sample-level top feature lists from Model Interpreter and super-feature mapping from Step I, we rank each sample’s top super-features by calculating narrative importance scores, and then use ultra-features to conduct deduplication (Section \ref{subsubsection:narrative_ranking}).
	\item
	Obtain information of top super-features for each sample: For each sample, we join the information of all super-features from Step II onto the top super-feature list from Step III.
	\item
	Construct top narratives for each sample: For each sample, we conduct narrative template imputation for each top super-feature (Section \ref{subsubsection:narrative_template_imputation}).
	\item
	(Optional) Construct top paragraphs for each sample: For each sample, we conduct narrative concatenation according to category name (Section \ref{subsubsection:narrative_concatenation}).
\end{enumerate}

\subsection{Narrative Exporter}

The generation of user-digestible narratives may not be the last step of a user-facing model explainer, instead the narratives should be further surfaced to various end user platforms such as sales/marketing intelligence platforms, Tableau dashboards and emails. Our solution is to incorporate an extra step called Narrative Exporter after the Narrative Generator, to unify the narrative surfacing process. Specifically, Narrative Exporter takes top narratives from Narrative Generator as its input, and converts them into a few specific formats of choice, such as html or email format. This step completes the end-to-end pipeline from machine learning platforms to end user platforms in CrystalCandle.

\section{Use Cases at LinkedIn}\label{sec:use_case}

LinkedIn leverages data to empower decision making in every area. One such area is sales, where data scientists built predictive machine learning models for account recommendation, covering the entire business landscape from customer acquisition to existing customer retention. Most of these predictive models are black-box models, making it challenging for data scientists to surface model outputs to sales teams in an intuitive way.

Furthermore, LinkedIn sales teams use multiple internal intelligence platforms. One typical platform, Merlin, aims to help sales representatives close deals faster by providing personalized and actionable sales recommendations/alerts. Before CrystalCandle, all these sales recommendations were rule-based. A typical example of rule-based recommendations is based on exploratory data analysis: Recommend the jobs upsell opportunity if views per job increased more than 10\%, or the number of job posts increased more than 20\% in the past month. As we can see, these rule-based recommendations were neither very accurate as model predictions nor scalable in their generation process.

CrystalCandle has assisted LinkedIn data scientists in converting machine intelligence from business predictive models into sales recommendations on platforms such as Merlin, where LinkedIn data scientists are typically both model owners and CrystalCandle users, while sales teams are the end users applying CrystalCandle's narrative insights to their work. One typical example of CrystalCandle-based sales recommendations on Merlin is with jobs upsell alerts. As introduced in Section \ref{sec:introduction}, the jobs upsell model predicts how likely each account is to purchase more job slots.

\begin{figure*}[t]
	\centering
	\begin{tabular}{cc}
		\includegraphics[width=0.302\textwidth]{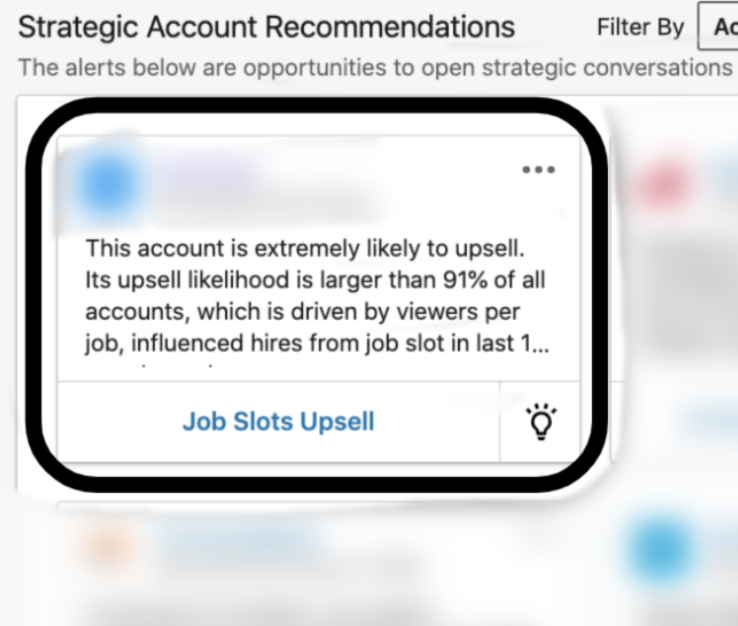} & \includegraphics[width=0.548\textwidth]{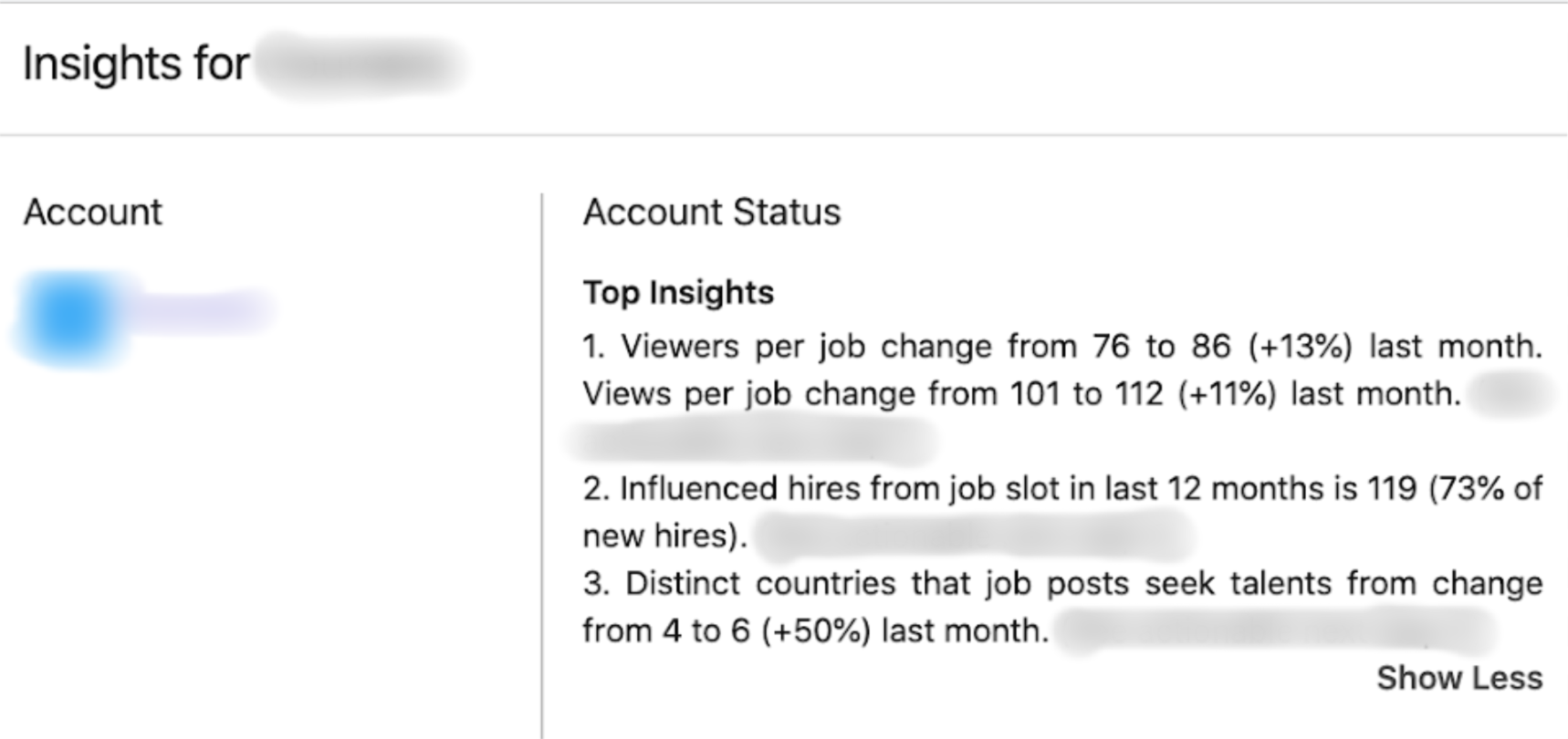} \\
		(a)	& (b) \\
	\end{tabular}
	\caption{Jobs Upsell Alerts on Merlin.} \label{fig:merlin_page}
\end{figure*}

Figure \ref{fig:merlin_page} shows how the jobs upsell alerts appear on Merlin. When a sales representative logs into Merlin, a list of account alerts including jobs upsell alerts are displayed on the Merlin homepage (Figure \ref{fig:merlin_page}(a)). On the summary page of the account, we see a sentence describing its propensity score. To learn more about the underlying reasons behind its recommendation, sales representatives can click the “Job Slots Upsell” button which will direct them to the account status page with more account details (Figure \ref{fig:merlin_page}(b)). In the Account Status section, top narrative insights are listed, e.g., both \texttt{viewers per job} and \texttt{distinct countries that job posts seek talents from} largely increased in the last month, which serve as strong signals of upsell propensity.

Besides Merlin, CrystalCandle-based sales recommendations have also been surfaced onto other sales platforms for different audiences and use cases with the help of Narrative Exporter. By the end of 2020, six CrystalCandle-based sales recommendations across four lines of LinkedIn business - Talent Solutions (LTS), Marketing Solutions (LMS), Sales Solutions (LSS) and Learning Solutions (LLS) have been on-boarded onto four internal sales intelligence platforms, which have been surfaced to more than five thousand sales representatives overseeing more than three million accounts.

\subsection{Evaluation Results}

To understand how helpful CrystalCandle-based sales recommendations are to sales representatives, we turned to qualitative and quantitative evaluations of CrystalCandle performance.

In our qualitative evaluation, we collected feedback from sales representatives via questionnaires, interviews, and other feedback channels. Similar approaches have been proposed in \cite{reiter2019natural} and \cite{sovrano2019difference}, where the authors argued that “subjective satisfaction is the only reasonable metrics to evaluate success in explanation”. We have conducted a survey within a small group of sales representatives on the helpfulness of CrystalCandle-based sales recommendations (ratings from 1 - not helpful at all to 5 - couldn't do my job without them). Ten responses have been received with average satisfaction rating of 3.5 (standard error 0.4). We have also collected positive feedback from a broader group of sales representatives, which we summarized into three main points (A collection of feedback in the original can be found in Section \ref{feedback_from_sales_representatives} in the Appendix):
\begin{enumerate}
	\item
	Top narrative insights are clear to understand and effectively help sales representatives build trust in the account recommendations. These narrative insights bring important metrics to their attention, and prompt them to work on accounts that they may have not considered otherwise.
	\item
	CrystalCandle-based sales recommendations serve as a comprehensive information center. Sales representatives appreciate that the top narrative insights are consolidated all in one place, to save their time of gathering information from different sources.
	\item
	CrystalCandle-based sales recommendations provide a directional guidance for next steps. The top narrative insights allow sales representatives to act strategically, e.g., prepare customer-specific conversations.
\end{enumerate}

\noindent Another way to evaluate the performance of CrystalCandle-based sales recommendations is via quantitative evaluation, which we conducted in two phases:
\begin{enumerate}
	\item
	Phase I: Compare the adoption rate between CrystalCandle-based recommendations and rule-based recommendations. Table \ref{table:interaction_rate} shows the interaction rates on all the CrystalCandle-based and rule-based Merlin Alerts across sales representatives in LTS and LLS respectively in the same 3-month time period, where the interaction rate is defined as \# clicks / \# impressions. CrystalCandle-based alerts have a significantly higher interaction rate than rule-based alerts, indicating that sales representatives are more engaged with CrystalCandle-based alerts. We note that potential confounding factors in this comparison may exist, e.g., the novelty of new alerts may lead to increased interactions. To address this novelty effect, we started our measurements of \# impressions and \# clicks one month after the new alerts launch date, in the hope that most of the sales representatives have already been familiarized with them. We have also extended the time period of measurements to three months to further reduce this potential novelty effect.
	\item
	Phase II: Identify the differences with/without CrystalCandle-based recommendations via A/B testing. In the A/B testing design, for each sales representative, we randomly split his/her account book into treatment/control groups, and we show the CrystalCandle-based recommendations to all the eligible accounts in the treatment group only. We then compare key metrics between the treatment and control groups, e.g., upsell rate and revenue for upsell recommendations (upsell rate = \# successful upsell opportunities / \# sales opportunites created), and churn rate and revenue lost for churn risk notifications. Table \ref{table:ab_testing} shows the A/B testing results of jobs upsell alerts and recruiters upsell alerts after a 3-month testing period,  where ``recruiters” is another LinkedIn product. We observe boosts in both key metrics: upsell rate and average spend per account, indicating that the CrystalCandle-based sales recommendations work effectively in driving the right sales decisions and bringing in revenue to the company. We note that the A/B testing period of recruiters upsell alerts was during the COVID-19 crisis, where severe hiring freezes likely negatively impacted the market of recruiters products \cite{kimbrough2020hiring}, leading to lower than expected numbers of created opportunities and upsell opportunities in Table \ref{table:ab_testing}. As a result, the corresponding upsell rate lift of recruiters product is not statistically significant, and its average spend per account possesses a relatively large standard error. 
\end{enumerate}

\begin{table*}[h]
	\caption{Interaction rate (\# clicks / \# impressions) of Merlin Alerts, CrystalCandle-based vs rule-based (standard error in parathesis).}\label{table:interaction_rate}
	\centering
	\begin{tabular}{|c|c|c|c|}
		\hline
		& & \textbf{LLS} & \textbf{LTS} \\
		\hline
		\textbf{CrystalCandle-based Alerts} & \# impressions & 694 & 7,188 \\
		& \# clicks & 41 & 167 \\
		& interaction rate (s.e.) & 5.9\% (0.9\%) & 2.3\% (0.2\%) \\
		\hline
		\textbf{Rule-based Alerts} & \# impressions & 1,031 & 5,445 \\
		& \# clicks & 25 & 91 \\
		& interaction rate (s.e.) & 2.4\% (0.5\%) & 1.7\% (0.2\%) \\
		\hline
		\textbf{Interaction Rate Lift} & lift (\%) & +141\% & +39\% \\
		& lift (p-value) & $<$0.001*** & 0.012** \\
		\hline
	\end{tabular}
	\\ ~\\
	\caption{A/B testing results of jobs upsell alerts and recruiters upsell alerts (standard error in parathesis).}\label{table:ab_testing}
	\centering
	\begin{tabular}{|c|c|c|c|}
		\hline
		\textbf{Alerts Type}
		& \textbf{Treatment/}
		& \textbf{Upsell}
		& \textbf{Avg Spend}\\
		& \textbf{Control}
		& \textbf{Rate (s.e.)}
		& \textbf{Per Account (s.e.)}\\
		\hline
		Jobs Upsell & With Alerts & 17.0\% (1.1\%) & \$65,733 (\$4,703) \\
		Alerts & No Alerts & 14.0\% (1.7\%) & \$32,626 (\$4,500) \\
		& Lift (\%) & +21\% & +101\% \\
		& Lift (p-value) & 0.084* & 0.074* \\
		\hline
		Recruiters & With Alerts & 9.6\% (2.7\%) & \$25,863 (\$14,209) \\
		Upsell Alerts & No Alerts & 5.9\% (2.2\%) & \$2,908 (\$1,236) \\
		& Lift (\%) & +63\% & +789\% \\
		& Lift (p-value) & 0.214 & 0.071* \\
		\hline
	\end{tabular}
	\\ ~\\
	\raggedright \small{*significance codes:  0 `***' 0.01 `**' 0.05 `*' 0.1 `' 1}\\
	\raggedright \small{**treatment/control split: 70/30 in jobs upsell alerts, 50/50 in recruiters upsell alerts}
\end{table*}

\subsection{Lessons from Deployment}

We list key lessons we have learned from the deployment of CrystalCandle at LinkedIn:
\begin{enumerate}
	\item
	Initial feedback we received after the launch of our first-ever CrystalCandle-based sales recommendations - jobs upsell alerts, was from sales representatives who found some narratives confusing. For example, one narrative was ``Distinct countries of job posts changed value”, which was identified as a vague statement: Did this mean ``distinct countries where job post viewers come from” or ``distinct countries where job posts seek talents from”?
	
	To resolve the above issue, we worked with data scientists who built sales predictive models to host multiple working sessions with sales representatives. During working sessions, we walked through the top narrative insights with sales representatives, asked them about meaningfulness of feature descriptions and whether more granular information was needed, and then revised the feature descriptions and narrative templates in Insights Design accordingly. For example, when we identified which description was correct for ``distinct countries of job posts”, we updated the corresponding super-feature in Insights Design, which solved the above issue. Overall, we found it useful to iterate through several rounds of improvements and feedback from our end users.
	
	\item
	Making the narrative templates reusable is strongly recommended to CrystalCandle users, in order to save efforts in template construction and maintenance.
	A certain group of super-features can be narrated by one template, while others cannot. For example, the template \texttt{value\_change} in Table \ref{table:narrative_templates} can only be applied to a super-feature with its original-feature pair in chronological order, such as \texttt{job\_view\_s3} and \texttt{job\_view\_s4} in Table \ref{table:insight_type} (\texttt{s3} and \texttt{s4} stand for ``two months ago” and ``last month” respectively). However, this template cannot be used for the original-feature pair \texttt{job\_qty} and \texttt{job\_dprice\_usd}, and instead we need to design a new template for it, such as template \texttt{quantity} in Table \ref{table:narrative_templates}.
	
	Therefore, when constructing narrative templates, we recommend that CrystalCandle users start with templates that accommodate as many of the super-features as possible, and then make new templates as needed until the rest of the super-features are covered. We found that if a narrative template is generalizable enough, it can also be used in other model interpretation use cases as well. In the Merlin use case, for example, over 80\% of narrative templates used in jobs upsell alerts were reused in recruiters upsell alerts, which shortened the preparation time of recruiters upsell alerts by more than half. Frequently used narrative templates in CrystalCandle can be stored for potential future usage.
\end{enumerate}

\section{Limitations and Future Work}\label{sec:limitations}

Here, we list several limitations of CrystalCandle and discuss future work:
\begin{itemize}
	\item
	CrystalCandle only supports supervised machine learning models whose input features are in tabular data format. Future work will aim to extend CrystalCandle to support a broader range of supervised learning models such as image classification and natural language processing models, as well as other types of models including unsupervised learning, semi-supervised learning and time series models.
	\item
	The Insights Design input, including the feature info file and narrative templates, is mostly manually created. We plan to investigate ways to auto-generate parts of Insights Design to further reduce manual efforts from CrystalCandle users.
	\item
	Translation of narrative templates into code is manually conducted. As future work, we will try to automate this process by identifying symbols and characters in narrative templates and converting them into appropriate code automatically.
\end{itemize}

\section{Conclusion}\label{sec:conclusion}

In recent years, requests from end users of predictive models of understandable model outputs have become widespread, motivating the development of user-facing model explainers. In this paper, we proposed CrystalCandle, a novel user-facing model interpretation and narrative generation tool, which produces user-digestible narrative insights and reveals the rationale behind predictive models. The evaluation results in LinkedIn’s use cases demonstrate that the narrative insights produced by CrystalCandle boost the adoption rate of model recommendations and improve key metrics such as revenue.

\section*{Acknowledgements}

We would like to express our special thanks to our colleagues at LinkedIn that put this work together, including Saad Eddin Al Orjany, Harry Shah, Yu Liu, Fangfang Tan, Jiang Zhu, Jimmy Wong, Jessica Li, Jiaxing Huang, Kunal Chopra, Durgam Vahia, Suvendu Jena, Ying Zhou, Rodrigo Aramayo, William Ernster, Eric Anderson, Nisha Rao, Angel Tramontin, Zean Ng, Ishita Shah, Juanyan Li, Rachit Arora, Tiger Zhang, Wei Di, Sean Huang, Burcu Baran, Yingxi Yu, Sofus Macskassy, Rahul Todkar and Ya Xu. We particularly thank Justin Dyer, Ryan Rogers and Adrian Rivera Cardoso for their helpful comments and feedback.

\newpage 
\bibliographystyle{ieeetr}
\bibliography{intellige_reference}

\newpage
\appendix

\section{Appendix}\label{sec:appendix}

\subsection{Translate Narrative Templates into Code}\label{subsec:translate_template}

We use the narrative template \texttt{value\_change} as an example to show how we can translate it into Scala code. This example can be easily generalized to other narrative templates and programming languages. Just to recap, the narrative template \texttt{value\_change} is: “\{super\_name\} changed from \{prev\_value\} to \{current\_value\} (\{percent\_change\}\%) in the last month”.

To calculate the extra insight item \texttt{percent\_change}, we can first build a helper function \texttt{changePercent} in Scala:
\begin{lstlisting}[language=Scala]
	def changePercent(current_value: Double, previous_value: Double): String = {
		val change_percent = 
		(current_value, previous_value) match {
			case (0, 0) => 0.0
			case (a, b) => (a / b - 1) * 100
		}
		change_percent match {
			case x if x.isInfinity => " "
			case x if x >= 0 => " (+" ++ change_percent ++ "%) "
			case _ => " (" ++ change_percent ++ "%) "
		}
	}
\end{lstlisting}

\noindent This helper function is used to first calculate the percent change from \texttt{prev\_value} to \texttt{current\_value} as \texttt{change\_percent}, and then convert \texttt{change\_percent} into a more user-friendly format: Empty if \texttt{change\_percent} is infinite (e.g., changed from 0 to 4), (+X\%) if \texttt{change\_percent} is positive (e.g., changed from 2 to 4 (+100\%)), and (-X\%) if \texttt{change\_percent} is negative (e.g., changed from 4 to 2 (-50\%)).

With the help of this helper function, we then build the Scala function \texttt{valueChangeInsight} for the narrative template \texttt{value\_change} which enables feature value imputation:
\begin{lstlisting}[language=Scala]
	def valueChangeInsight(super_name: String, insight_item: Map[String, Double]): String = {
		val change_percent_desc: String = 
		changePercent(insight_item("current_value"),  
		insight_item("prev_value"))
		(super_name.capitalize ++ " changed from "
		++ insight_item("prev_value") ++ " to "
		++ insight_item("current_value")
		++ change_percent_desc ++ " in the last month.")
	}
\end{lstlisting}

\noindent This Scala function has two inputs:
\begin{enumerate}
	\item
	\texttt{super\_name}: The name of super-feature. E.g., \texttt{views per job}.
	\item
	\texttt{insight\_item}: A Scala Map which maps all the insight items in the narrative template to their corresponding feature values for each sample. E.g., for customer A, \texttt{insight\_item = Map("prev\_value" -> 100, "current\_value" -> 150)}.
\end{enumerate}

\noindent In practice, we can conduct narrative template imputation by simply calling this Scala function. For example, we can run the following Scala code if we want to construct narrative insight of super-feature \texttt{views per job} for customer A:
\begin{lstlisting}[language=Scala]
	val narrativeInsightA = valueChangeInsight("views per job", Map("prev_value" -> 100, "current_value" -> 150))
\end{lstlisting}
\noindent The output will be “Views per job changed from 100 to 150 (+50\%) in the last month”.

\subsection{Additional Columns in Feature Info File}\label{subsec:additional_column}

We briefly introduce three additional columns Insight Threshold, Insight Weight and Source in Feature Info File. A sample Feature Info File for the jobs upsell use case with these three columns included is shown in Table \ref{table:feature_info_file} (Due to space limitations, we omit columns that already exist in Table \ref{table:feature_hierarchy} and \ref{table:insight_type}). CrystalCandle users can work with model owners to fill in these three columns, and adjust their values based on feedback collected from end users:
\begin{table}[h]
	\caption{Feature Info File for selected features from jobs upsell model.}\label{table:feature_info_file}
	\centering
	\begin{tabular}{|c|c|c|c|c|}
		\hline
		\textbf{Original-} & \textbf{$\cdots$} & \textbf{Insight} & \textbf{Insight} & \textbf{Source} \\
		\textbf{Feature} & & \textbf{Threshold} & \textbf{Weight} & \\
		\hline
		job\_qty & $\cdots$ & & 0.8 & model \\
		\hline
		job\_dprice\_usd &$\cdots$ & & 0.8 & user\\
		\hline
		job\_view\_s3 & $\cdots$ & percent\_ & 1 & model \\
		& & change$>$10 & & \\
		\hline
		job\_view\_s4 & $\cdots$ & percent\_ & 1 & model \\
		& & change$>$10 & & \\
		\hline
		job\_viewer\_s3 & $\cdots$ & percent\_ & 1 & model \\
		& & change$>$10 & & \\
		\hline
		job\_viewer\_s4 & $\cdots$ & percent\_ & 1 & model \\
		& & change$>$10 & & \\
		\hline
		job\_applicant\_s3 & $\cdots$ & percent\_ & 1 & model \\
		& & change$>$5 & & \\
		\hline
		job\_applicant\_s4 & $\cdots$ & percent\_ & 1 & model \\
		& & change$>$5 & & \\
		\hline
	\end{tabular}
\end{table}

\begin{itemize}
	\item
	\textbf{Insight threshold:} This threshold can be used to filter out those narratives not meeting certain criteria, so that the remaining narratives can be more relevant to end users. For example, the insight threshold for super-feature \texttt{views per job} is \texttt{percent\_change>10}, thus only the narratives with the increment of job views larger than 10(\%) will be shown to end users.
	\item
	\textbf{Insight weight:} This weight can be used to make adjustments to the narrative ranking. It takes values between 0 and 1 (default is 1), and is multiplied with the feature importance score from Model Interpreter to determine the final importance score for each original-feature. The motivation is to incorporate domain knowledge into narrative ranking: If we believe some original-features are predictive in modeling but not that informative to end users, we can lower their insight weights to prevent prioritizing their corresponding narratives.
	\item
	\textbf{Source:} Sometimes additional features from external data sources can also be used in narrative construction (together with model features). These additional features are usually in the formats incompatible with the predictive models (e.g., name, date and url), however, they can help make the generated narratives more informative. For example, a narrative can be “This customer spent 15 hours browsing websites last week, with the most visited website \textit{xyz.com}”, where 15 is a model feature, and \textit{xyz.com} is an additional feature which can not be fed into the model directly. We can set “source” to be \texttt{model} or \texttt{user} to specify the source of each original-feature. Note that one narrative cannot be constructed by using only the additional features, i.e., the additional features must be paired with at least one model features under the same super-feature, to make sure that a valid narrative importance score can be assigned for narrative ranking.
\end{itemize}

\newpage

\subsection{Narrative Ranking Example}\label{subsec:narrative_ranking_example}

Table \ref{table:narrative_ranking} shows one example of conducting narrative ranking and deduplication in jobs upsell use case for customer A. We set $K=1$ in narrative deduplication.

\begin{table}[h]
	\caption{Example of narrative ranking and deduplication in jobs upsell use case.} \label{table:narrative_ranking}
	\centering
	\begin{tabular}{|c|c|c|c|}
		\hline
		\textbf{Original-} & \textbf{Super-} & \textbf{Ultra-} & \textbf{Feature} \\
		\textbf{Feature} & \textbf{Feature} & \textbf{Feature} & \textbf{Imp. Score} \\
		\hline
		job\_qty & job slots & job slots & 0.3 \\
		\hline
		job\_dprice\_usd & job slots & job slots & 0.4 \\
		\hline
		job\_view\_s3 & views & job view & 0.2 \\
		& per job & & \\
		\hline
		job\_view\_s4 & views & job view & 0.6 \\
		& per job & & \\
		\hline
		job\_viewer\_s3 & viewers & job view & 0.3 \\
		& per job & & \\
		\hline
		job\_viewer\_s4 & viewers & job view & 0.2 \\
		& per job & & \\
		\hline
	\end{tabular}
	~\\~\\
	$\Downarrow$ \textbf{Narrative Importance Score Calculation}\\
	~\\
	\begin{tabular}{|c|c|c|c|}
		\hline
		\textbf{Super-} & \textbf{Ultra-} & \textbf{Narrative} & \textbf{Narrative} \\
		\textbf{Feature} & \textbf{Feature} & & \textbf{Imp. Score} \\
		\hline
		job slots & job slots & Purchased 30 & 0.4 \\
		& & job slots ... & \\
		\hline
		views & job view & Views per job & 0.6 \\
		per job & & changed ... & \\
		\hline
		viewers & job view & Viewers per job & 0.3 \\
		per job & & changed ... & \\
		\hline
	\end{tabular}
	~\\~\\
	$\Downarrow$ \textbf{Narrative Deduplication}\\
	~\\
	\begin{tabular}{|c|c|c|}
		\hline
		\textbf{Top Ultra-} & \textbf{Top Narrative} & \textbf{Narrative} \\
		\textbf{Feature} & & \textbf{Imp. Score} \\
		\hline
		job view & Views per job changed ... & 0.6 \\
		\hline
		job slots & Purchased 30 job slots ... & 0.4 \\
		\hline
	\end{tabular}
\end{table}

\subsection{Feedback from Sales Representatives}\label{feedback_from_sales_representatives}

We list several comments from sales representatives in their original words:
\begin{itemize}
	\item
	``These are awesome. I LOVE that you’ve called out the \% of employees. All of these are SUPER helpful. The top insights are clear and concise. It would have taken me a lot of time to find all of that information, so I love that it is all laid out. It also prompts me to think more strategically, which I love."
	\item
	``Most accounts were on my radar, but perhaps not for the reasons highlighted in the insights—so calling that out provides a new perspective on potential entry point into an account/ways to actively engage with relevant insight/reason."
	\item
	``This SAR is clear to understand and is valuable by bringing important metrics to my attention. As I dig into SARs a bit more, time will tell which insights are most valuable to me."
	\item
	``As someone new to LI, this is incredibly helpful. It points me into the right direction and allows me to take action quickly and in a way that correlates to an activity we are seeing in Sales Navigator."
	\item
	``I love that the insights are consolidated all in one place, versus needing to run different reports in Merlin to gather the same information. I love the piece that highlights how many days reps are logging onto LinkedIn and searching on the platform."
	\item
	``Yes it's very valuable and useful for conversations. It helps with next steps and the insights will help it lots of different ways to tell a story to a customer depending on where that conversation is at or to help prospect in."
\end{itemize}

\end{document}